\newif\ifmypreprint 

\ifmypreprint
  \documentclass[review]{elsarticle}
\else
  \documentclass[3p,twocolumn]{elsarticle}
\fi

\usepackage{amsmath} 
\usepackage{amsfonts}
\usepackage{mathtools}
\usepackage{mymacros}
\usepackage{lineno,hyperref}
\usepackage[english]{babel} 
\modulolinenumbers[5]
\journal{Journal of Systems Architecture}

\hypersetup{draft}

\begin{document}
\newcommand{\newsection}{\acresetall \newpage}

\newcommand{\pdataw}{\text{data}_{\text{size}}^{(\text{W})}}
\newcommand{\pdatai}{\text{data}_{\text{size}}^{(\text{I})}}
\newcommand{\pdmainputbw}{\text{DMA}_{\text{BW}}^{(\text{input})}}
\newcommand{\pdmaoutputbw}{\text{DMA}_{\text{BW}}^{(\text{output})}}
\newcommand{\ppen}{\text{PE}_{\text{N}}}
\newcommand{\ppebw}{\text{PE}_{\text{BW}\{\times\}}}
\newcommand{\ppeii}{\text{PE}_{\text{II}}}
\newcommand{\pclbcolsize}{\text{CLB}_{\text{size}}^{(\text{col})}}
\newcommand{\pclbshiftsize}{\text{CLB}_{\text{size}}^{(\text{shift})}}
\newcommand{\pclbrown}{\text{CLB}_{\text{N}}^{(\text{rows})}}
\newcommand{\pdboutbw}{\text{DB}_{\text{BW}\{\times\}}^{(\text{output})}}
\newcommand{\pwbsize}{\text{WB}_{\text{size}}^{(\text{buffer})}} 
\newcommand{\ppoolsize}{\text{pool}_{\text{size}}^{(\text{depth})}}

\newcommand{\cwbbw}{\text{WB}_{\text{BW}\{\times\}}}
\newcommand{\cwbbwbit}{\text{WB}_{\text{BW}}}
\newcommand{\pkernelsn}{\text{kernels}_{\text{N}}^{(\mathbb{R}^{3 \times 3 \times 512 + 1})}}

\begin{frontmatter}

\title{A scalable and efficient convolutional neural network accelerator using HLS for a System on Chip design}

\author[mymainaddress]{Kim Bjerge\corref{mycorrespondingauthor}}
\cortext[mycorrespondingauthor]{Corresponding author}
\ead{kbe@ase.au.dk}
\author[mysecondaryaddress]{Jonathan Horsted Schougaard}
\author[mysecondaryaddress]{Daniel Ejnar Larsen}
\address[mymainaddress]{School of Engineering, Aarhus University, Finlandsgade 22, 8200 Aarhus N, Denmark}
\address[mysecondaryaddress]{Department of Engineering, Aarhus University, Finlandsgade 22, 8200 Aarhus N, Denmark}

\begin{abstract}


This paper presents a configurable Convolutional Neural Network Accelerator (CNNA) for a System on Chip design (SoC).
The goal was to accelerate inference of different deep learning networks on an embedded SoC platform.
The presented CNNA has a scalable architecture which uses High Level Synthesis (HLS) and SystemC for the hardware accelerator. 
It is able to accelerate any Convolutional Neural Network (CNN) exported from Python and supports a combination of convolutional, max-pooling, and fully connected layers. 
A training method with fixed-point quantized weights is proposed and presented in the paper.
The CNNA is template-based, enabling it to scale for different targets of the Xilinx Zynq platform. 
This approach enables design space exploration, which makes it possible to explore several configurations of the CNNA during C- and RTL-simulation, 
fitting it to the desired platform and model. 
The CNN VGG16 was used to test the solution on a Xilinx Ultra96 board using PYNQ.
The result gave a high level of accuracy in training with an auto-scaled fixed-point Q2.14 format compared to a similar floating-point model.
It was able to perform inference in $2.0$ seconds, while having an average power consumption of $2.63$ W, 
which corresponds to a power efficiency of $6.0$ GOPS/W. 

\end{abstract}

\begin{keyword}
    \texttt System On Chip \sep FPGA \sep High Level Synthesis \sep Convolutional Neural Network \sep PYNQ 
\end{keyword}

\end{frontmatter}

\ifmypreprint
  \linenumbers
\fi

\section{Introduction}

In recent years, deep learning with Convolutional Neural Networks (CNNs) has been applied in many different fields 
such as image classification~\cite{Krizhevsky2017},\cite{Simonyan2015}, object detection~\cite{Redmon2016},\cite{Ren2017} and recognition~\cite{He2018}.  
In most cases, state-of-the-art CNN models run on a server in the cloud.
However, with the increase of Internet of Things (IoT), there is a demand for embedding the deep neural networks into mobile edge computing.
This is especially true for computer vision systems, where the amount of collected data is high and analyses of images must be carried out in real-time.
 
As CNNs continue to be applied to increasingly complex problems, low throughput, latency and energy efficiency 
present challenges on embedded devices with Central Processing Units (CPUs) or Graphical Processing Units (GPUs). 
Due to several attractive features, Field Programming Gate Arrays (FPGAs) present promising platforms for Hardware (HW) acceleration of CNNs as reported in~\cite{Shawahna2019},\cite{Mittal2014},\cite{Mittal2018},\cite{Ding2019}. 
CNNs that are optimized for fixed-point data or use binary neural networks achieve even better performance~\cite{Hubara2016},\cite{Blott2018},\cite{Nurvitadhi2017},\cite{BNN-Accel}.  
In general, FPGAs provide higher performance than CPUs and have a better energy efficiency than both CPUs and GPUs.

Historically, the long design time and need for HW experts have limited the use of FPGAs. 
Here, the high-level synthesis tools have enabled automatic compilation from imperative high-level programs 
to low-level specifications in a Hardware Description Language (HDL)~\cite{Nane2016}. 
It is, however, still a challenge to accelerate large-scale CNNs~\cite{Li2016} on a FPGA, 
since model parameters typically require far more memory than the on-chip capacity of the FPGAs. 
Another challenge is to find an optimal configuration for a given HW accelerator design due to the long design time.

The scope of our work is to develop a generic and flexible architecture, 
which can accelerate the inference of CNNs on a Multi-Processor System on Chip design (MPSoC). 
It presents the design of the HW/SW architecture, i.e. the programmable logic that will reside in the FPGA fabric and the design of the software.
The architecture is generic so that it can accept major CNNs such as AlexNet~\cite{Krizhevsky2012} and VGG16~\cite{Simonyan2015}, 
which can be exported from a deep learning framework such as Keras~\cite{Keras}. 
It is developed in the PYNQ~\cite{Stornaiuolo2018} framework using Python and SystemC~\cite{systemCstandard} in order to create a generic template based HW accelerator. 
To find the optimal design, this study uses a SystemC-based simulation to explore the design space of the optimal configuration parameters of the Convolutional Neural Network Accelerator(CNNA). 
The design model is translated to a HDL specification using High Level Synthesis (HLS).
Our paper discusses the precision, speed and power consumption of the accelerator as well as the fixed-point retraining of the CNN.

\subsection{Related work}

In this section, the current state-of-the-art HW-based CNNAs that inspired the architecture presented in this paper will be discussed. 

The Microsoft model~\cite{Ovtcharov2015} is an architecture developed by Microsoft for accelerating CNN for a cloud server solution with several FPGA-cards. 
The architecture uses a top-level controller to control the data-flow with a PCI memory interface. 
It has multiple input buffers, one kernel weight buffer, a large array of Processing Element Arrays (PEAs) and lastly, a data redistribution block. 
It uses a Direct Memory Access (DMA) channel to load data in from PC memory to the buffers.
On the FPGA it uses PEA blocks to perform dot product calculations of the values in the input buffer and the weight buffer. 
The result of the dot product is saved into the next input buffer. 

ZynqNet~\cite{zynqnet} is based on the architecture of the Microsoft model. 
However, it focuses on making it work for both training and inference. 
It is built for a System on Chip (SoC) design instead of a server solution. 
The proposed solution seems promising, although it appears to have a few bottlenecks due to a purely C-based HLS implementation of the solution.
It uses a Circular Line Buffer (CLB) for input data handling and a memory-mapped master interface to get data from the main memory, 
i.e. weights and input data are transferred using the memory-mapped interface. 

FINN-R\cite{Blott2018} is an end-to-end deep-learning framework for fast exploration of Quantized Neural Networks (QNNs). 
It is a framework built upon the FINN accelerator~\cite{Umuroglu2017} which is a QNN built for FPGA. 
The FINN-R consists of a cascade of multiple layer accelerators that are optimized for a pipelined architecture. 
This design reduces the transfer of data between the main memory and the accelerators. 
The difficult part is to balance the layered accelerators in order to prevent bottlenecks or resource waste. 
However, the framework does not solve the problem of different throughput for each layer.
FINN-R optimizes the generated HW using HLS, allowing fast exploration of QNN to create the perfect accelerator for a specific FPGA target. 

To accelerate and develop CNNs on reconfigurable HW (FPGAs), a survey of the current State-of-the-art toolflows was published in 2018 by Venieris et al.~\cite{Venieris2018}. 
The survey compares the performance of: fpgaConvNet~\cite{Venieris2016,Venieris2019}, D{\footnotesize NN}W{\footnotesize EAVER}~\cite{Sharma2016}, Angel-Eye~\cite{Guo2018}, DeepBurning~\cite{Wang2016} and Caffeine~\cite{Zhang2016}.
The listed toolflows covers mapping of the classic AlexNet and VGG16 to the Xilinx Zynq or UltraScale platforms.
The deep learning framework Caffe by Berkeley AI Research is the most widely supported front end for these State-of-the-art toolflows.  

The above-mentioned toolflows can be divided into two main categories of architectures: \emph{streaming architecture} 
and \emph{single computation engine}. 
FINN-R, fpgaConvNet and DeepBurning are in the category of \emph{streaming architectures}. 
This chained and pipelined architecture can be achieved high performance, but the optimal HW design needs to be found and synthesized for each specific CNN.
The \emph{single computation engine}, on the other hand, executes CNN layers sequentially, which means that the same HW engine is able to handle many different CNNs.
The engine is controlled from SW, and data must be moved from CPU memory to the on-chip FPGA memory during processing of the CNN layers.  
The advantage of this approach is that the same HW can be used for several CNN architectures without reconfiguration.

Tunable parameters must be optimized for the available resources of the FPGA devices, which is the case for D{\footnotesize NN}W{\footnotesize EAVER}, Angel-Eye and Caffeine. 
Angel-Eye uses a compiler to translate the input CNN to custom instructions. 
A similar approach is used by D{\footnotesize NN}W{\footnotesize EAVER}, which utilizes a macro dataflow instruction set architecture and supports FPGAs from both Xilinx and Altera.    
Of the above-mentioned toolflows, only fpgaConvNet supports special layers with irregular dataflow including the inception, residual and dense blocks that are required for the newest deep neural networks such as ResNet~\cite{Huang2016}, DenseNet~\cite{Huang2017}, InceptionNet and GooglLeNet~\cite{Zeng2016}. 
      
The \emph{single computation engines} build their architectures around a PEA with a buffer for handling input data, which could be a CLB or a row buffer. 
A streaming design, such as FINN-R uses the output to feed the next accelerator.
Consequently, the \emph{streaming architecture} has a large memory to cache layered outputs directly to the next input buffer so that the data are ready for the next CNN layer. 
However, due to limited internal memory, this approach is not feasible for all FPGAs.
Therefore, there is a need for reloading the input data from the main memory. An example of this is the Microsoft model. 
Other architectures use the main memory to cache the data between layers. FINN-R and fpgaConvNet, does this for each block of layers. 

The CNN developed in this work has some elements in common with the solutions presented above. 
It uses the main memory to store data between layers and uses the \emph{single computation engine} approach. 
In addition, the architecture is built around a PEA with two buffering systems: one for the weights and one for input image data, the latter of which uses a CLB.  
The above architectures are very similar, but the major difference lies in the details of the CLB, which enables efficient pipelining and data allignment.
 
The CNN architecture in our work supports any input size and layer depth, stride, zero padding and windows size.  
It makes the accelerator more flexible and enables it to run nearly any CNN model that uses convolution, pooling and fully connected layers. 
It can be used with most CNNs during run-time inference without the need for re-compiling. 
The accelerator is developed to work with PYNQ~\cite{pynq},\cite{Stornaiuolo2018} and uses an Application Programming Interface (API) similar to Keras~\cite{Keras}.
In summary, this paper makes the following contributions.

\begin{itemize}

\item We present a generic CNN architecture consisting of a \emph{single computation engine} with five core elements (weight buffer, data buffer, PEA, pooling block and output handler) to perform FPGA acceleration of CNN models.

\item Stitching is used for convolutional layers that are too large to execute in a single processing pass and is used to split complex convolutions into sub convolutions.  

\item Dynamic auto scaling is used during training to minimize the accuracy between the floating point and the quantized fixed point accelerator.  

\item A template based SystemC design with an executable model is proposed for design space exploration.
The template model is synthesized to an Xilinx IP core with HLS and controlled from the host CPU using the PYNQ framework and Python.     

\end{itemize}

\section{Design methods}

In this section, we will briefly describe the design methods and concepts used as a basis for designing 
and implementing the architecture for the CNNA.

In our work, SystemC is used with the design flow described in~\cite[ch. 1]{ug902},\cite{synthesisSystemC}.
It is an efficient way in which an IP can be written and verified using HLS. 
SystemC is able to model the HW structure and concurrency in a more optimal manner than pure C and C++.
It is an IEEE standard (IEEEStd 1666-2011)~\cite{systemCstandard}, based on a C++ template library made for HW/SW co-design.

The use of templates makes the IP core configurable and portable to explore different solutions and platforms, whereas custom designs are less flexible.
It is much faster to recompile and simulate a template-based IP core than to write a custom IP that may be more optimal.
By use of SystemC the desired HW architecture can be controlled and designed via modules with parallel clocked threads.
With HLS directives it is possible to control the synthesized threads and achieve a desired unroll, pipeline and iteration interval of
the synthesized RTL code. 

PYNQ (Productiviy for Zynq)~\cite{pynqBook} is an open-source framework for creating applications on a ZYNQ MPSoC.
The system design in this work is based on PYNQ for controlling the IP directly from Python. 
This framework is divided into three layers: application, SW and HW. 

The application layer, which hosts the user-code, is  described in~\cite[ch. 22]{pynqBook}. 
This is usually Python code in the form of a Jupyter notebook that runs on the ARM CPU inside the Zynq MPSoC. 
The middle layer in the PYNQ framework is the SW layer. 
This layer contains the Python libraries and the interaction with the IP inside the FPGA through the OS drivers. 
Several drivers are provided through the PYNQ libraries for interacting with the IP. 
The interface is called an overlay and is used to program the FPGA and manage the IP.
The last HW layer in the PYNQ framework is the \texttt{bit}-file programmed into the FPGA. 
The interaction between the SW layer and the HW layer is done using DMA or memory-mapped interfaces. 

\section{System architecture}

The SoC design consists of three main elements: the FPGA (i.e. the Programming Logic (PL)) the dual core CPU and memory (i.e. Dynamic Random Access Memory (DRAM)). 
The goal is to run a CNN consisting of convolutional, max-pooling and fully connected layers computed in the same IP core inside the FPGA logic. 
The responsibility of the CPU is to let the user control the HW acceleration so that the IP core is processing CNN layers in correct sequential order. 
~\Figref{ibd-CNNA-system_v2.png} shows that the system uses DMA to transfer data and weights between the CPU and IP core accelerator. 
The CPU controls the DMA data block transfer and converts the memory interface to the streaming interface of the IP core. 

\myfig{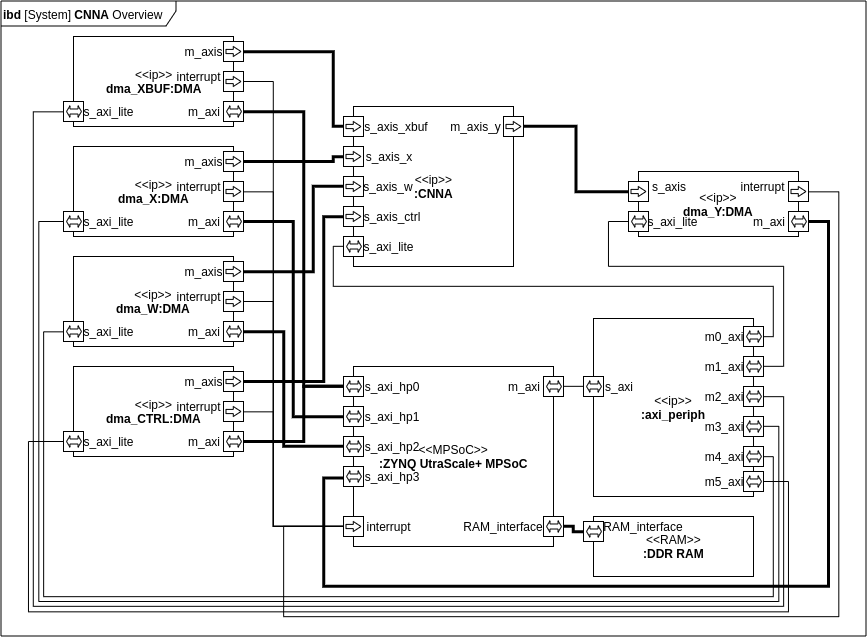}{1.00}
   {Block diagram of the system architecture covering CPU (Zynq Ultrascale+ MPSoC), memory (DDR RAM), HW IP core accelerator (CNNA), five DMAs for inputs (X), outputs (Y), weights (W), control (CTRL) and splits (XBUF).}

The system interacts in different manners depending on which scenario it needs to execute. 
There are three main scenarios: \emph{preprocessing}, \emph{initialization} and \emph{inference}. 

\paragraph*{Preprocessing}
The first scenario converts the weights to fixed-point and realigns and scales the weights so that they are ready for the system to use.
Preprocessing also calculates parameters such as layer output size and layers to be split which can be done offline on any computer.
The weights are transformed from floating-point to fixed-point representation in the chosen format, and aligned and rounded off correctly, 
as described later.
Finally, the weights are saved in an h5-file, which is the standard format for storing weights in Keras~\cite{Keras}, 
and can be transferred to the HW target.

\paragraph*{Initialization}
The HW target needs to be configured and initialized for a particular fixed-point resolution by using the synthesized bit-file of the optimized CNNA.
The bit-file contains the CNNA IP core and interconnection setup to the CPU for the specified HW target.  
This is done by using a specification of the model in the form of a JSON-file and an h5-file containing the weights, 
which are already realigned and quantized in the preprocessing scenario. 
It starts by calculating the buffer size and getting the properties of the loaded CNNA. 
When this is done, the SW allocates the needed resources and readies the SW for inference by allocating the buffers for each layer in the CNN. 

\paragraph*{Inference}
When using the system, predicting an image will be the most commonly used task. 
This task is shown in the sequence diagram in~\figref{sd-CNNA-system-pred.png}. 
Here, the user calls the method predict, which returns the predicted class of the image when the inference is done. 
The image, which is the parameter to the method predict, is stored internally in a contiguous array, 
i.e. an array which can be used by the PYNQ DMA library. 
Depending on the CNN, several layers are executed in the correct order, i.e. convolution, pooling or fully connected layer.
All parameters controlling the CNN execution are sent at the start of the predict method.

\myfig{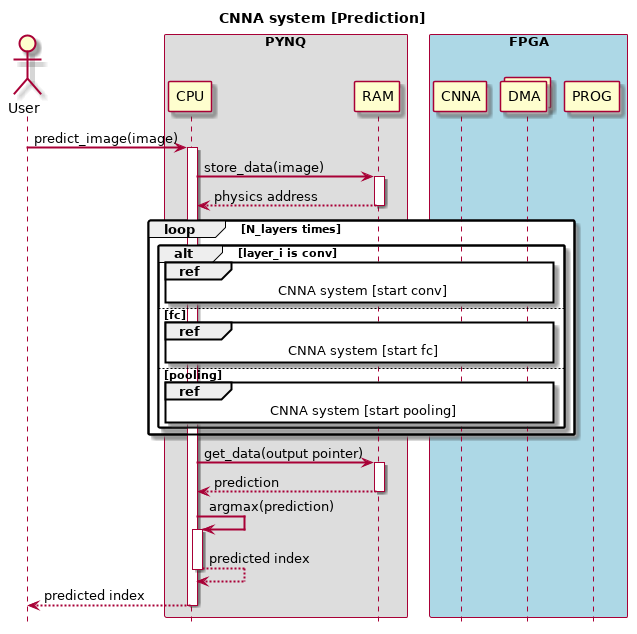}{1.00}
   {Sequence diagram of the system of interaction between SW control (PYNQ) and HW accelerator (FPGA) during inference.}

The convolution is done by the CPU initiating four different tasks in parallel. 
It sets up the data transfer for the input control data CTRL, the input data X, the output Y and XBUF. 
Each of these data transfers are handled by the DMA, which streams the content of the buffer from DRAM to the CNNA. 
The fully connected layer is executed similarly to both pooling and convolution. 
It starts four different DMAs: one for each of the input data X, the weights W, the output Y and the configuration though the CTRL. 

Two interfaces are used. 
The streaming interfaced used for the DMA is implemented with a functional deterministic guaranty 
so that no race condition can happen, which makes the entire IP very stable. 
The AXI streaming interface, is used for transmitting the data between the CPU and IP.
The other AXI-lite~\cite{UG761} interface, which is a memory-map, is only used for reading status registers. 

\subsection{Software control and stitching}
\label{sec:swArchCovnLayer}

Some convolutional layers are too large to be processed as a single CNNA iteration. 
This means that they are split into several sub-convolutions. 
However, the result is returned from the IP core with the depth first and thus needs to be stitched together with the later result. 
This is done through the IP core which has a DMA-channel (XBUF) for this purpose, as shown in~\figref{ibd-CNNA-system_v2.png}. 
An example of stitching can be seen in~\figref{bufferSplit.png}. 
The shown example illustrates the output of two pixels, a and b, from a convolution with a depth that needs to be split. 

\myfig{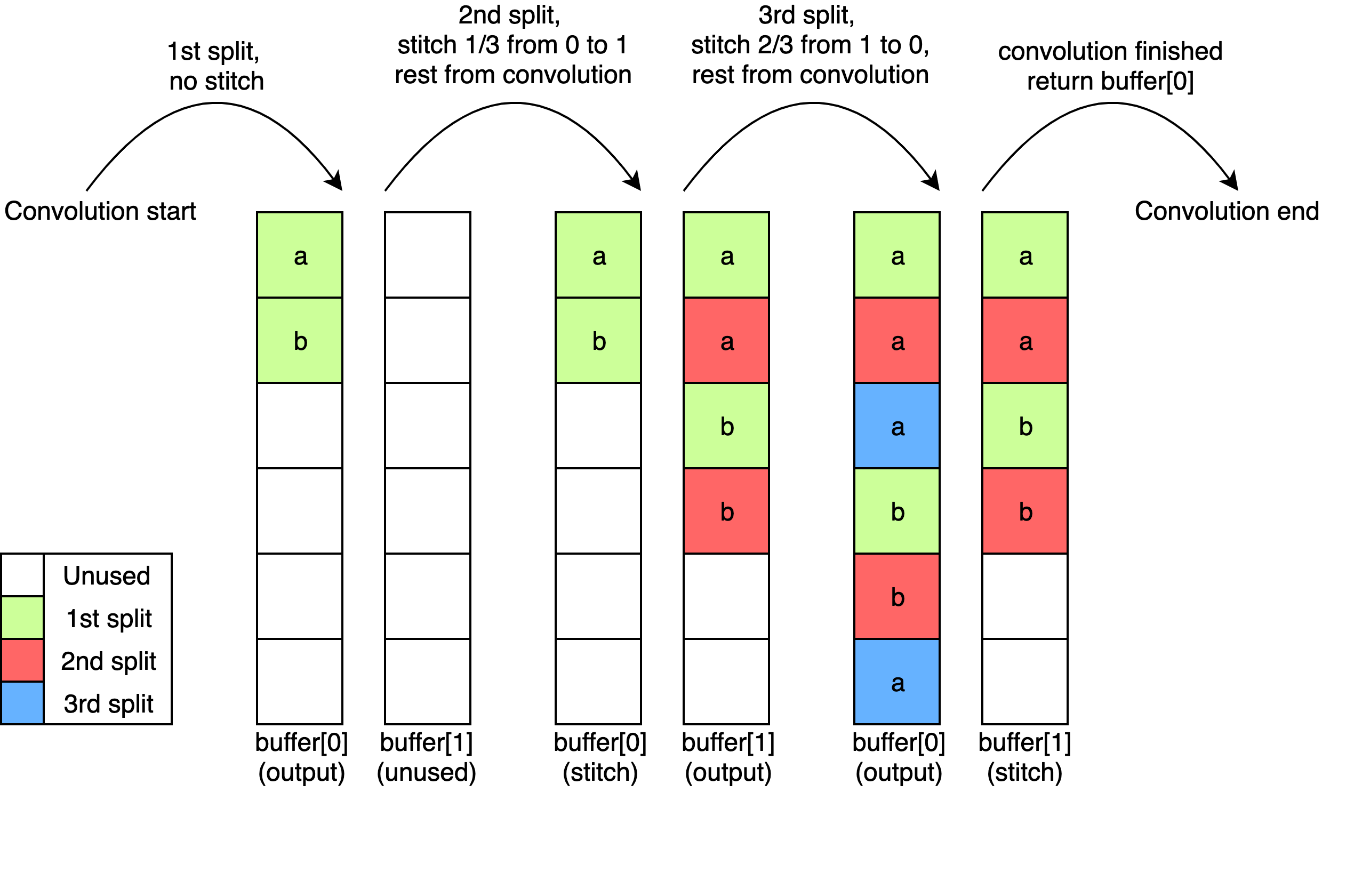}{1.00}
   {Example of buffer stitching with a split of three shown for a single pixel.}
	
The stitching is done by using two equally sized buffers, which both have a size equal to the expected output size.
The size in this example is six. 
The first convolution only uses the first buffer as the output, and pixels a and b are both updated by the DMA. 
However, only the first third of the depth is calculated in the first convolution. 
The second convolution calculates the next third of the output. 
However, these outputs need to be stitched in between the previous outputs. 
This is done by using the output of the first convolution as a stitch buffer. 
The IP core is informed to use the first part of each pixel and appends the result to each pixel depth-wise. 
The result of this stitching is sent to the output buffer[1]. 
The third convolution takes two thirds from the stitch buffer and the last third from the output for each pixel. 
The output of the stitched convolution is in the buffer, which is the one used as output in the last stitching.

However, the fully connected layers can also be too large to be processed at once, in which case the splits are handled differently. 
A fully connected layer generates a single value per output. 
The buffer will be filled with values from the first split when it runs the first time. 
The second time it runs it will get the next outputs, which must then to be placed after the first split in the buffer. 
All the splits need to have an adjusted number of bytes to ensure that the correct amount of data is received. 
When all the splits have been processed the result is in the same buffer. 
This means that the fully connected layer only needs a single buffer, contrary to the convolution layers, which needs two.

\subsection{CNN hardware accelerator}

\myfig{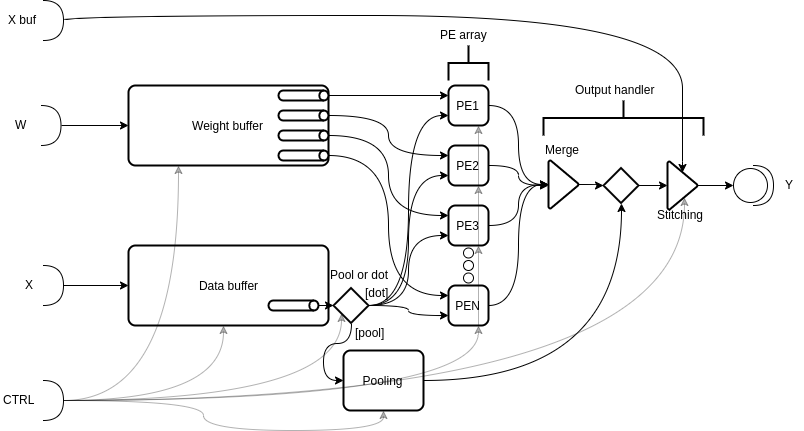}{1.00}
      {Illustration of the CNNA architecture with five streaming buffer inputs for control, stitching, weights and data as well as one result output buffer. 
			 A number of Processing Elements (PE) are used to accelerate the pooling and multiplications for each neuron in the network. 
			 The CNNA will be executed several times and typical once for each layer during inference.}
			
\Figref{CNNA_overview.png} shows the architecture overview of the main elements of the CNNA. 
The CNNA works as an accelerator for a single layer at a time. 
This means that the accelerator needs to be reconfigured for each layer, which is done through the streaming interface \emph{CTRL}.

The streaming interface \emph{W} is used to load the weights, which can consist of multiple kernels, 
and cache them in the weight buffer. This means that they can be used multiple times without reloading from DRAM. 
The streaming interface \emph{X} is used to stream in the input data, which can either be an image or the output from the previous layer. 
\emph{X buf} is an interface that is used when a convolutional layer is split into several splits, which need to be stitched together correctly. 
The last streaming interface is \emph{Y}, which streams out the output values of the operation.

The accelerator is built for three different operations: convolution, pooling and fully connected layers. 
During convolution acceleration, it uses the weight buffer, the data buffer and the PEA. 
The pooling operation is done by using only the data buffer and the pooling block. 
When executing the fully connected layers, the weight buffer and the data buffer are simply set to forward the data directly to the PEA, 
thus generating a dot product of the two.
The following sections describe the five core elements comprising the design of the CNNA.

\subsubsection{Weight buffer}

The weight buffer is used for caching the weight kernels. 
This caching is necessary, because the convolution requires the kernels to be used once for each output pixel. 

An illustration of the weight buffer module can be seen in~\figref{wb_align.png}, which shows the modules inside of it.
The Iteration Interval (II) and Bandwidth (BW) of the weight input package are changed during resizing as illustrated in the figure.
It shows how the resize module changes the BW from input BW, BW$^{(\text{in})}$ by a factor of resize$_\text{factor}$. 
It also splits the raw package into smaller packages. The realign module splits the raw package into smaller packages. 
The splitter separates the data stream into $N$ different stream buffers, each of which has the same BW as the resized BW, BW$^{(\text{resize})}$. 
Each stream buffer sends the kernel to a Processing Element (PE) $X$ times.

\myfig{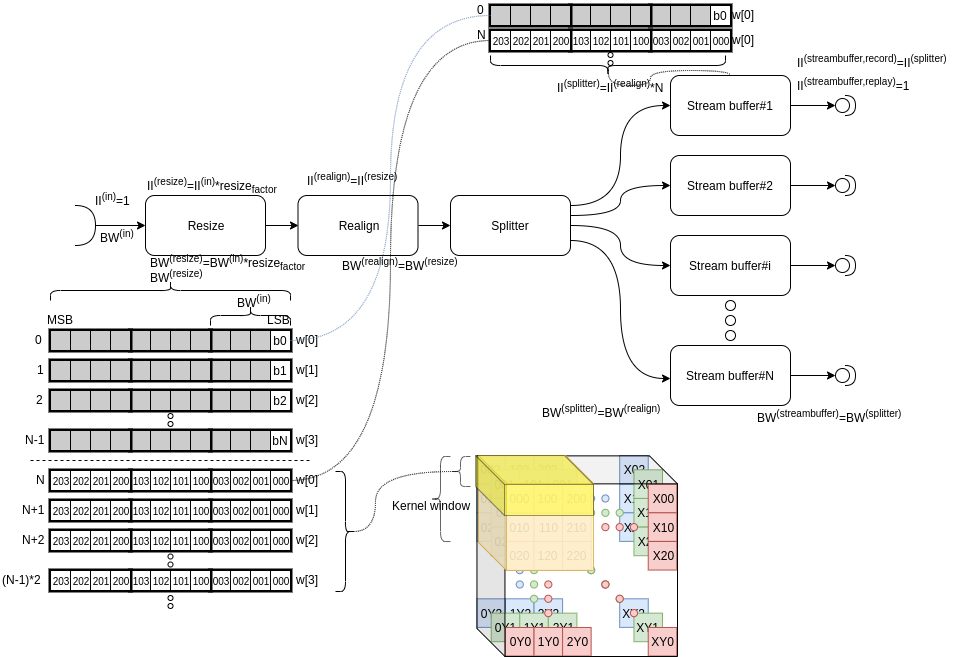}{1.00}
   {Weight buffer with alignment illustrates how kernel weight data are aligned so that a specific kernel is placed in the correct spot. 
	The illustration shows how the weight data of kernel 0 are sent to stream buffer 1. 
	The II and BW of the weight input package are changed by a factor of resize$_\text{factor}$.
	The yellow part of the image cube shows which part of the kernel is sent.}

The realignment in the weight buffer is complicated. Firstly, this is due to the bias value, which uses a complete package. 
Secondly, it is complicated because the kernels need to match the order in which the three-dimensional window comes from the data buffer, 
i.e. have the same positions and depths as the data buffer.
In~\figref{wb_align.png}, the first package contains the bias values transferred to the weight buffer. 
It shows that this single value only uses a complete resized package. 
This is followed by N other bias packages. After all bias packages are sent the weight packaged are sent. 
In this example, the weight packages contain a $3 \times 3 \times 4$ window. 
The stream buffers receive the values one after the other.

\subsubsection{Data buffer}
The data buffer is used to handle the input data stream and create the windows on which the convolution or pooling is carried out.

An image typically consists of three channels, RGB, which can be visualized as a three-dimensional cube. 
Such a three-dimensional image is illustrated in~\figref{clb_v2.png}. 
The image is stored in raster order, i.e. first, pixel (0,0) channel 0, then channel 1 of the same pixel followed by the last channel. 
This is followed by the same three channels for the pixel one row down, which means 
that the Z-axis is processed first, then the Y-axis second and the X-axis last. 
Raster order is the order in which the image data is streamed to the CNNA.

\myfig*{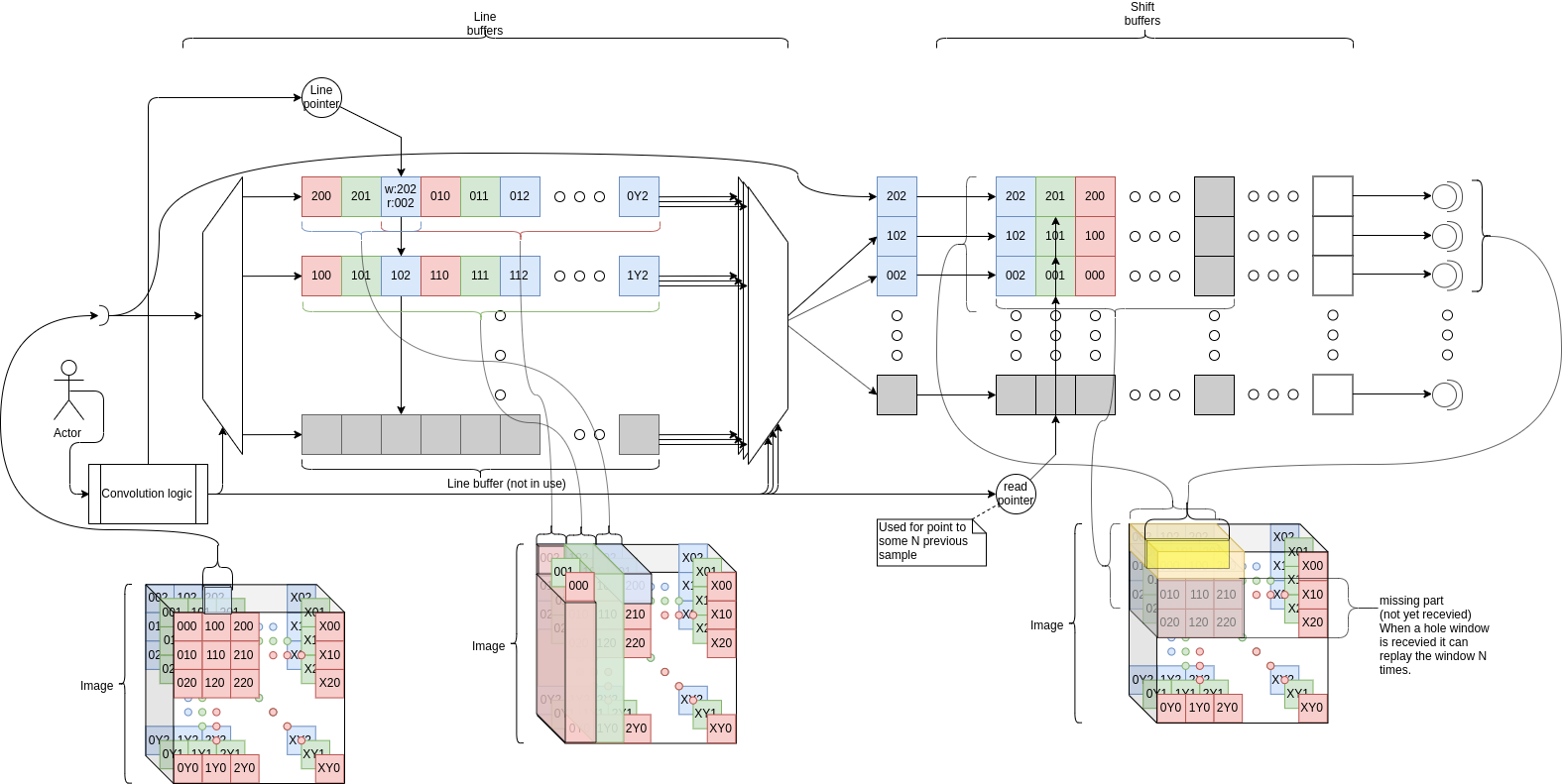}{1.00}
   {An illustration of the flow of data through the CLB. It consists of two parts: a line buffer for storing $N_{lines}$ previous lines and a shift buffer for storing $N_{pixels}$ previous pixel for each line. The leftmost image cube illustrates a single pixel $202$, which is written for the line buffer. The middle cube illustrates which data is saved in which line buffer and how the new line replaces the first line. The rightmost cube illustrates what data is in the shift buffers. The missing part illustrates how much more data is needed from the line buffers before it has a complete window in the shift buffer. The read pointer on the shift buffers is used for getting the N previous samples and for generating the output from the shift buffers.}

The CLB can be considered the brain of the CNNA, because it allows 
it to increase the BW and removes the need to realign the input data for each layer.
The parameters that the actor can set through the control interface are:

\begin{itemize}
	\item \emph{Row size} — The row size of the input image, i.e. the Y-axis length. It is assumed that the image is quadratic.
	\item \emph{Depth} — The depth is N-channels of the input image, i.e. the Z-axis of the image. This should be dividable by the BW.
	\item \emph{Stride} — Stride is a convolution parameter.
	\item \emph{Window size} — The size of the window. If the window size is 3, the real window size would be $ 3 \times 3 \times \text{depth} $. This is also a convolution parameter. 
	\item \emph{Zero pad} — A convolution parameter setting the size of the zero padding around the image.
	\item \emph{Replay} — How many times the CLB should resend a single window.
\end{itemize} 

After setting up the CLB with the parameters through the control interface, the image data can flow into the CLB.
The CLB consists of two parts: a line buffer for storing $N_{lines}$ previous lines and a shift buffer for storing $N_{pixels}$ 
previous pixel for each line. These parts are explained in detail below:

\paragraph*{Line buffers}
The first module in the CLB, where the image data is ordered and stored is the line buffer. 
This module streams one row of the image with all channels at the same time. 
The number of line buffers is equal to the maximum window size minus one, $N_\text{line buffer} = window_\text{size} - 1 $. 
This is because only the N previous lines are needed to construct a window. 
The illustration of the data flow of the line buffer in~\figref{clb_v2.png} 
shows that the $N-1$ previous lines are stored inside the line buffer and sent out individually. 
This means that the BW increases with a factor of $window_\text{size}$. 
It is also indicated that the buffer is stored circularly.  
This is handled by the pointer, which can be seen in~\figref{clb_v2.png}. 
This pointer will increase each time a new input is received. 
After receiving a whole line, the line buffers will rotate, i.e. the first line will be moved to the back the second line will be pushed forward and the pointer will be reset. 
This is done by multiplexing logic in the implemented design.

\paragraph*{Shift buffers} 
After the line buffers, the data reaches the shift buffers. 
These buffers are used for getting the N previous pixels from each line, i.e. having all the pixels needed for a convolution window, 
as shown in~\figref{clb_v2.png}.
The shift buffers have another important function as well. 
They replay the window for the convolution if there are not enough PEs to run all the dot products in the convolution operation at once.
The shift buffers are on-chip RAM-based shift buffers and consist of two pointers.
The write pointer is essentially controlled by counting up whenever data is written 
and moving it back to start of the shift buffer when the end has been reached.
The read pointer, however, is controlled by logic, which tells the shift buffer that it needs the N previous samples. 
This will be handled by the shift buffer, which also calculates its new positions.

\subsubsection{Processing element array}
The heart of the CNNA is the PEA. 
Each PE performs HW acceleration of a dot product with a small range of activation functions, e.g. linear or ReLU~\cite{relu.DBLP}. 
The PE operation can be written as shown in equation~\ref{eq:pe_math_equal}, which is a dot product of the two equal length vectors $\vec{x}$ and $\vec{w}$.

\begin{equation}\label{eq:pe_math_equal}
	PE(\vec{x},\vec{w}) = f\biggl(\sum_{i=0}^{N-1}( x_i \cdot w_i)\biggr)
\end{equation}

Each PE receives data frames in pairs from the weight buffer and the data buffer, i.e. one from each. 
The acceleration of the PE is done by running the multiplications in parallel and totaling the results afterward, as illustrated in~\figref{PE.png}.
This data is dotted together and followed by the activation function. 

\myfig{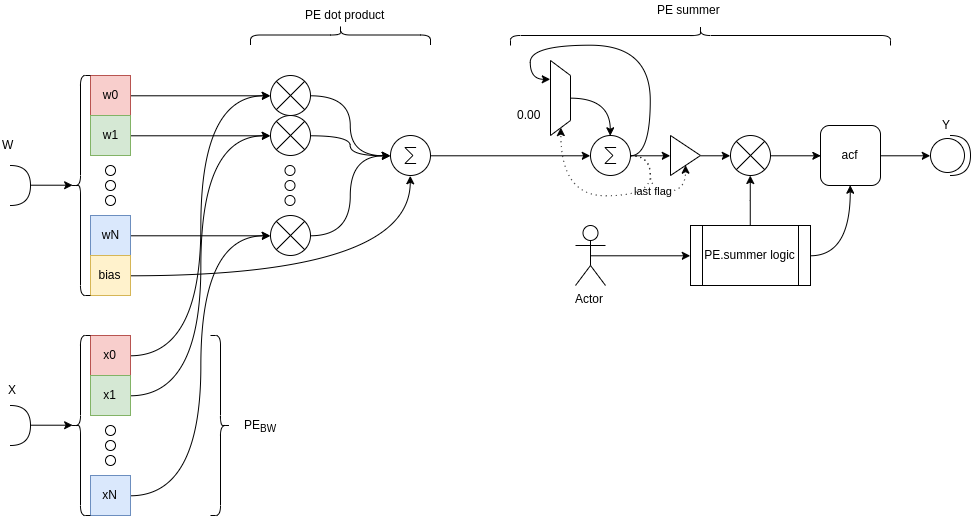}{1.00}
   {Illustration of the PE design. It consists of a parallel multiplier array followed by a summing tree and lastly an accumulator, 
	  scale and the activation function logic. The PE dot product and summer are executed in a parallel and pipeline order.}

\Figref{PE.png} shows how the PE has two inputs: \emph{W}, the weight input, and \emph{X}, the data input. 
When the PE has received a frame on both \emph{W} and \emph{X}, the data frames are dotted together and the bias is added to the result.
The result is forwarded to the next part, which is the pipelined PE summer. 
This part accumulates the result, which it has received from the PE dot product. 
It will keep on accumulating until it receives the last flag. 
When this happens, it will multiply the accumulated value by a factor set by the actor, i.e. the control interface, 
and apply the activation, which is also set by the actor through the control interface. 
Lastly, it is streamed out through the port \emph{Y}, and the accumulated result is reset. 
The hole PE is synthesized with a II of one, which means that new inputs can be processed in each clock cycle.
HLS try to solve this problem with a summer tree or a cascade of Multiply-Accumulates (MACs) processed in a long pipeline.  

\subsubsection{Pooling}
The pooling element is used to accelerate the pooling operation. 
It gets its input from the data buffer and sends the output to the output handling part, 
thus bypassing the PEA, which is not used in pooling. 

The reason for placing the pooling operator inside the CNNA is reuse of the CLB hardware.
The pooling accelerator receives its input directly from the CLB, and the output from the pooling goes directly to the output. 

\myfig{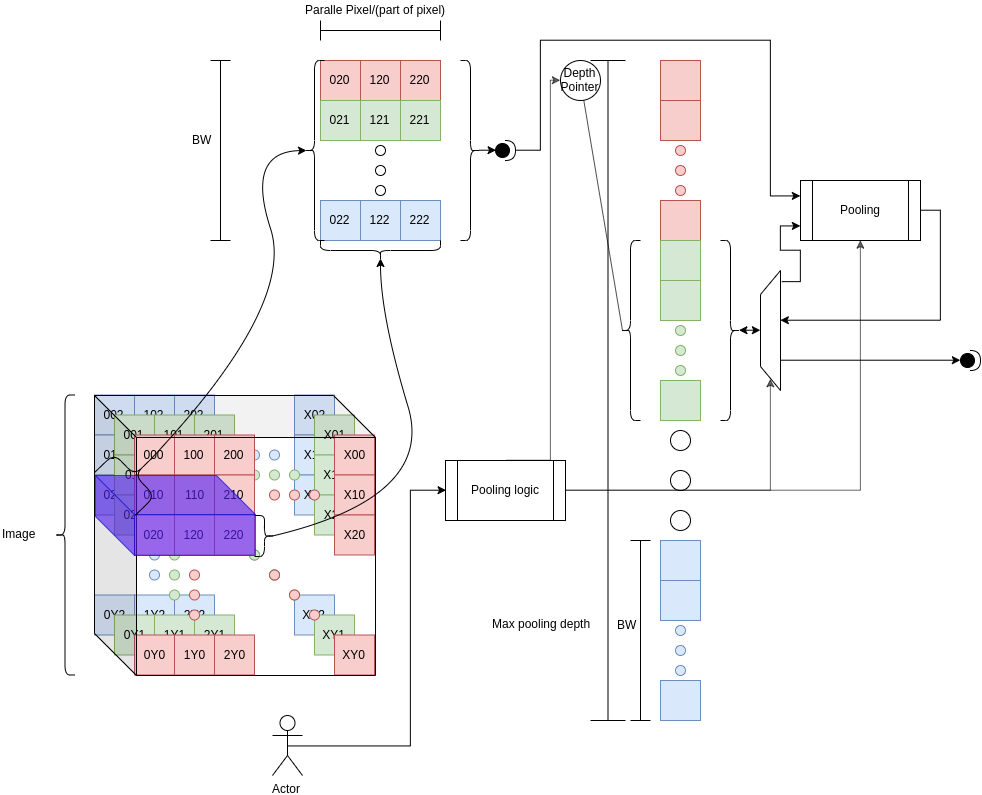}{1.00}
   {Illustration of the logic of the pooling accelerator. The data is received as a small slice (the purple cube), which is the output of the CLB. It compares the input data with the current pooling. If it is the first value, it saves it. After pooling, logic is executed. If it is max-pooling it checks if the input is larger than the stored value, and stores the largest one.  When a whole window has been run through the accelerator, it will start streaming out the calculated pixels. These steps are repeated for all the windows created by the CLB.}

When looking at~\figref{pooling_cnna.png}, it can be seen that the pooling block consists of logic for handling the pooling operation, 
e.g. max-pooling, and RAM for buffering a single pixel. 
The pooling logic is controlled by the actor and is used for setting the depth of the current image and the size of the window, 
e.g. $2 \times 2 \times \text{depth}$ or $3 \times 3 \times \text{depth} $. 
The last parameter controls what type of pooling operator should be run, i.e. max-, min- or average-pooling.

\subsubsection{Output handler}
The output handling element plays a major role in getting the output of the CNNA into the correct shape 
and alignment before streaming it out through interface \emph{Y}. 
It merges the results from the PEA when it is used, and if the data needs to be stitched with \emph{X buf}, 
i.e. if a convolution operation has been split into more than one convolution and needs to have the old output interleaved into the new output. 
Splitting the convolution happens when too many kernels need to be stored in the weight buffer.
It also handles the output of a pooling operation, which simply means forwarding the output of the pooling element. 

\section{Training for fixed-point}
\label{sec:result:training}

To overcome the challenge of the CNNA using fixed-point values, an emulation of fixed-point needs to be made in order for the CNN to be trained and calculated correctly. 
This is mostly due to the large dynamic range of the weights.

This emulation is shown in equation~\ref{eq:float_2_fixed}, where $Q_{[I.F]}(x)$ 
is the fixed-point representation of $x$ in the fixed-point format Q[I].[F]~\cite{Oberstar}. 
Here, I is the number of integer bits and F is the number of factional bits.
First, the number $x$ is scaled up by $2^{F}$ and then rounded off to resolve the limited resolution of fixed-point numbers. 
This is followed by what is essentially a saturation of the number to the range of the fixed-point number, i.e. between  $-2^{I+F-1}$ and $2^{I+F-1}-1$. 
Lastly, the number is scaled down by the same factor it was scaled up by. 
This results in a value that can be interpreted correctly by the CNNA.

\begin{equation}\label{eq:float_2_fixed}
\begin{split}
Q_{[I.F]}(x) =\max\biggl(-2^{I+F-1},~  \min\bigl(2^{I+F-1}-1, \\
                round(x \cdot 2^{F})\bigr)\biggr) \cdot 2^{-F}
\end{split}
\end{equation}

\subsection{Quantized weights}

The weights are quantized as a constraint to the optimizer, which executes the backpropagation~\cite{backprog}. 
This constraint is set to quantize all weights after each update using equation \ref{eq:float_2_fixed}.
This results in the Stochastic Gradient Decent (SGD) update formula shown in equation~\ref{eq:quantize_freez}, where $Q_{[I.F]}(x)$ 
is the quantization function shown in equation~\ref{eq:float_2_fixed}, $W_{ij}^{(l,t=t-1)}$ 
is the previous weight, $W_{ij}^{(l,t=t)}$ is the new weight, and $\alpha$ is the learning rate.

\begin{equation}\label{eq:quantize_freez}
W_{ij}^{(l,t=t)} = Q_{[I.F]}(W_{ij}^{(l,t=t-1)} -\alpha \nabla_{W_{ij}}^{(l,t=t-1)})
\end{equation}

However, this introduces a problem, that makes the training freeze.
The cause of the problem is that the size of the update to the weights is too small to move from one quantized value to another.
The effect of a too-small update change can be seen in the example shown in equation~\ref{eq:quantize_freez_eg}. 
It is not possible to update a single weight in Q2.6 with a value smaller than the smallest quantized value, in this case $2^{-6} = 0.015625$.
The example shows a weight with value $1.671875$ being updated by a too-small value: $0.0015624$. 
Updating the quantized weight value does not result in a change, which causes the training to freeze.

\begin{equation}\label{eq:quantize_freez_eg}
W_{ij}^{(l)} = Q_{[2.6]}(1.671875 - 0.0015624) = 1.671875
\end{equation}

To solve this issue, an extra copy of the weights $W$ is saved so that the forward pass, i.e. inference, 
is calculated using the quantized weights, and the SGD is calculated using unquantized weights. 
This means that the weights do not get stuck between quantization steps. 
This is also known as lazy update SGD~\cite{lazySGD}.
In this way, the weights $W$ are saved and the quantized weights $WQ$ are used for the forward pass, 
which can be seen in equations \ref{eq:quantize_freez_2_1} and~\ref{eq:quantize_freez_2_2}.

\begin{equation}\label{eq:quantize_freez_2_1}
W_{ij}^{(l),t=\tau} = (W_{ij}^{(l),t=\tau-1} -\alpha (\nabla_{W}^{(l),t=\tau-1})_{ij})
\end{equation}

\begin{equation}\label{eq:quantize_freez_2_2}
WQ_{ij}^{(l),t=\tau} = Q_{[I.F]}(W_{ij}^{(l),t=\tau})
\end{equation}
 
By using these equations, the optimizer can train the CNN even though the changes are too small to be significant when quantized.

\subsection{Dynamic range scaling}

The small kernels in the first convolutional layers of the CNN VGG16 have large weights, i.e. close to $ 1 $ or $ -1 $, 
but the fully connected layers have very small weights that only use the lowest bits, even in Q2.14. 
This means that the CNN needs more fractional bits. 
However, this is possible to solve by dynamically scaling the weights and the output. 
This is carried out with integers in~\cite{Jacob_2018_CVPR}. 

The following will show how this can be carried out on fixed-point values as well.
It has been found that the dynamic range of each kernel is almost the same for each layer. 
This knowledge can be used to add scaling to each layer in order to change the dynamic range of the weights. 
For example, based on the given weights
\begin{equation*}
W = \begin{bmatrix}
0.11 & 0.024 & -0.30 \\ -0.05 & 0.002 & 0.1
\end{bmatrix}
\end{equation*}
and a fixed-point format Q[I].[F], which, for simplicity, is able to store a maximum value of 1, denoted $Q_{[I.F]}^{MAX}$, a scaling can be found. 
To find the scaling needed for a better dynamic range, equation~\ref{eq:dynamic_scale} can be used. 
This equation takes the absolute maximum absolute value of the weights and divides it by the maximum value of the fixed-point format. 
 
\begin{equation} \label{eq:dynamic_scale}
scale^{(l)} = \frac{ \max_{i}\biggl( |{W_{i}^{(l)}}| \biggr)}{Q_{[I.F]}^{MAX}} = |-0.30|= 0.30
\end{equation}

The scaled value of the weights can now be calculated as shown in equation~\ref{eq:dynamic_W_scale}, which divides the weights by $scale^{(l)}$. 
This shows that the maximal absolute value is now -1.

\begin{equation} \label{eq:dynamic_W_scale}
W_{scale}^{(l)} = \frac{W^{(l)}}{scale^{(l)}} = \begin{bmatrix}
0.367 & 0.08 & -1 \\ -0.167 & 0.00667 & 0.333
\end{bmatrix}
\end{equation}

Using this scale factor, the output of a layer is calculated as shown in equation \ref{eq:dynamic_z_scale}, which has an added multiplication of the quantized value of the scale factor, where $z_{scale}^{(l)}$ is the scaled output of layer $l$, $W_{scale}^{(l)}$ are the scaled weights, $\vec{a^{l-1}}$ is the output from the previous layer and $scale^{(l)}$ is the scale factor of the layer $l$. 

\begin{equation} \label{eq:dynamic_z_scale}
z_{scale}^{(l)} = Q_{[I.F]}(W_{scale}^{(l)} \cdot \vec{a^{l-1}}) \cdot  Q_{[I_{scale}].[F_{scale}]}(scale^{(l)})
\end{equation}

Because of the quantization, it cannot be guaranteed that the outputs are the same, but they should be very similar, i.e. $z^{(l)} \simeq z_{scale}^{(l)}$.  
The main difference between the scaled and unscaled version is that $z_{scale}^{(l)}$ is better suited for the bit range of the fixed-point format than $z^{(l)}$. 

\section{Design space exploration}
 
The template-based IP core written in SystemC has a number of parameters that must be selected in order to achieve an optimal solution.
The executable model of the IP core gives an approximate estimate of the time performance and FPGA resource usage.   
When optimizing the IP core for a FPGA, it can be an enormous task to generate a design and find the optimal design parameters. 
It takes approximately one hour to synthesize the HLS code to RTL code. 
If this was to be carried out for all possible combinations of parameters, it would take weeks, 
months or even years, since the architecture has such a large number of parameters, 
e.g. BW between modules, FIFO-depth, the number of PEs, etc.
The high level model of the IP in SystemC can be simulated faster than the RTL code by a factor of 50-200 times, depending on the size of the accelerator. 
It is possible to use a heuristic approach to find the optimal solutions for a certain fixed-point resolution 
constrained by the given target device by evaluating several simulated solutions.

\myfig{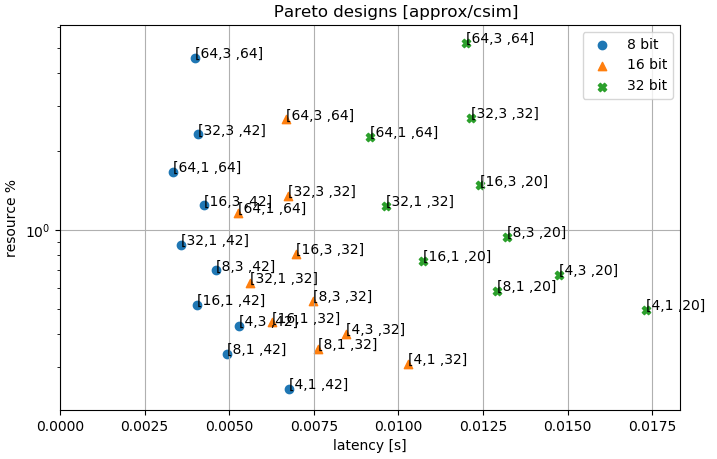}{1.00}
   {Results of the C-simulation of the combined test of all fixed-point candidates, showing the average resource usage of BRAM and DSP versus latency. The $\ppebw$ is set to $128$ for all solutions. The plotted text for a candidate is in the format $\biggl[\ppen,\pdboutbw, \pkernelsn\biggr]$. The candidates are split up in three groups of the word lengths ($\pdataw$) 8 bit, 16 bit and 32 bit versions. It took 2 minutes to create the estimate for one candidate solution.}
	
The design parameters are used for tuning the CNNA design in order to find a balance between precision, speed and resources.  
The CNNA tuning parameters used are as follows:

\paragraph*{$\pdataw$}
The word-length of the fixed-point data format in bits, i.e. I+F. Has an impact on precision.

\paragraph*{$\ppebw$}
The internal BW with an element size of $\pdataw$ used by the CNNA. 
	
\paragraph*{$\ppen$}
The number of PEs and the PEA are limited by the size of FPGA fabrics.

\paragraph*{$\pdboutbw$}
The output BW multiplier after the CLB. Normally this will be set at an equal value to $\pclbrown$, 
but can be set to a lower number in order to allow the PE to run with lower BW and potentially have a bigger PEA. 
The internal BW in the PE will be $ \pdboutbw \cdot \ppebw$, with an element size of $\pdataw$. 
The BW used inside the weight buffer is also equal to $ \pdboutbw \cdot \ppebw $.

\paragraph*{$\pkernelsn$}
Used to calculate
\begin{equation*}
\begin{split}
\pwbsize = \biggl(\frac{(3 \times 3 \times 512)}{\pdboutbw \cdot \ppebw} + \text{bias}_{\text{size}}\biggr) \\
           \cdot \pkernelsn
\end{split}
\end{equation*}
Here $(3 \times 3 \times 512)$ is chosen from the largest layer in the CNN, which, in this case, is the VGG16~\cite{Simonyan2015}.  
 	
The tuning of the CNNA can be expressed as a vector, $\vec{\beta}$, with 5 hyper-parameters, as shown in equation:
\begin{equation*}
\begin{split}
\vec{\beta} = \biggl[\pdataw,\ppebw,\ppen,\pdboutbw, \\
              \pkernelsn\biggr]
\end{split}
\end{equation*}
	
\begin{table*}[htbp]
  \centering
	\caption{Design space exploration of resource usage and latency of possible CNNA candidates using C- and RTL-simulation. 
	         A "$\div$" means RTL-simulation performed, but insufficient space on target platform (Ultra96). A "-" means RTL-simulation not performed.}
	\label{tab:dse:final}
	\begin{tabular}{|c|r|r|r|r|r|r|r|}
		\hline
		\multicolumn{1}{|c|}{\begin{tabular}[c]{@{}c@{}}Parameters \\ $\vec{\beta}$ \end{tabular}} & \multicolumn{1}{c|}{\begin{tabular}[c]{@{}c@{}}resource\\ average \%\end{tabular}} & \multicolumn{1}{c|}{DSPs} & \multicolumn{1}{c|}{\begin{tabular}[c]{@{}c@{}}DSPs\\ (RTL)\end{tabular}} & \multicolumn{1}{c|}{BRAMs} & \multicolumn{1}{c|}{\begin{tabular}[c]{@{}c@{}}BRAMs \\ (RTL)\end{tabular}} & \multicolumn{1}{c|}{latency$[ms]$} & \multicolumn{1}{c|}{\begin{tabular}[c]{@{}c@{}}latency$[ms]$\\  (RTL)\end{tabular}} \\ \hline
		$[8,128,8,3,42]$ & $ 70$ & $384$ & $359$ & $144$ & $249$ & $ 4.60 $ & $6.66$ \\ \hline
		$[8,128,8,1,42]$ & $ 34$ & $128$ & $125$ & $137$ & $185$ & $ 4.95 $ & $7.28$ \\ \hline
		$[8,128,16,3,42]$ & $ 124$ & $768$ & - & $152$ & - & $ 4.26 $ & - \\ \hline
		$[8,128,16,1,42]$ & $ 52$ & $256$ & $245$ & $139$ & $193$ & $ 4.03 $ & $6.66$ \\ \hline

		$[16,128,8,3,32]$ & $ 54$ & $192$ & $360$ & $233$ & $377$ & $ 7.47 $ & $8.40$ \\ \hline
		$[16,128,8,1,32]$ & $ 35$ & $64$ & - & $227$ & - & $ 7.61 $ & - \\ \hline
		$[16,128,16,3,32]$ & $ 81$ & $384$ & $\div$ & $239$ & $\div$ & $ 6.98 $ & $\div$ \\ \hline
		$[16,128,16,1,32]$ & $ 44$ & $128$ & $293$ & $229$ & $349$ & $ 6.27 $ & $8.52$ \\ \hline		
		
		$[32,128,8,3,20]$ & $ 94$ & $384$ & - & $355$ & - & $ 13.19 $ & - \\ \hline
		$[32,128,8,1,20]$ & $ 58$ & $128$ & $165$ & $351$ & $336$ & $ 12.92 $ & $14.53$ \\ \hline
		$[32,128,16,3,20]$ & $ 148$ & $768$ & - & $359$ & - & $ 12.42 $ & - \\ \hline
		$[32,128,16,1,20]$ & $ 76$ & $256$ & $325$ & $353$ & $408$ & $ 10.73 $ & $12.81$ \\ \hline
		
	\end{tabular}
\end{table*}

To measure the performance of the different CNNA configurations, a simulation was made. 
It consisted of five different elements: two pooling operations, two convolution operations and a single fully connected operation. 
They were executed individually but evaluated together.

When looking at the latency for the combined simulation test, i.e. the five simulations carried out consecutively after each other, 
the dominant candidates all have $\pdboutbw = 1$ regardless of word length (see \figref{pareto_designs_csim.png}). 
The figure shows that the faster the accelerator, the higher the number of PEs.

Two models were created of each configuration, one of which was done using C-simulation, 
i.e. a simulation that used the SystemC HLS code directly.
The other was a RTL-simulation, which used the RTL-code generated from the SystemC model for the most optimal solutions. 
The latter was clock cycle accurate and the execution time was precise.
	
Several candidates were identified and shown in greater detail in table \ref{tab:dse:final}. 
The table shows the number of Digital Signal Processing slices (DSPs) and BRAM used, as well as the total latency for C- and RTL-simulation. 
Some candidates marked with a "-" used more resources than were available on the tested target platform. 

The candidates with lowest latency were synthesized and tested using RTL-simulation, which simulates the real HDL-code generated. 
This also gives a more precise resource usage, which only differs slightly from the ones estimated using C-simulation. 
The execution time is also shown and is slightly higher (approx. 2ms) than the estimated value.
On average, the compilation and C-simulation time took 2 minutes for each solution. 
The HLS synthesizatione and RTL-simulation took 1-7 hours. 

The optimal parameters were found using two different fixed-point formats ($\pdataw$): Q2.14 and Q2.6, 
i.e. a word length of 16 bits and 8 bits, respectively. 
These were chosen because of the area constraints of the FPGA on the Xilinx Ultra96 board~\cite{ultra96}.
However 32-bits would have been possible with a larger FPGA. 

Finally, three different configurations of the CNNA were chosen for the final test of the system, 
one of which used 16-bit fixed-point format Q2.14, while the two others used 8-bit fixed-point format Q2.6.

\section{Results and discussion}

The dataset DETECT~\cite{detect} was used to verify the system. 
This dataset consisted of 29 classes of micro-invertebrates suspended in alcohol.
Only the first five classes were used in the first test, while the second test used all 29 classes.
Cifar-100~\cite{Krizhevsky2009} and ImageNet~\cite{ILSVRC15} were used for comparison with other common datasets and to validate the results.
The training was carried out over the span of 100 epochs.

The CNN used was VGG16~\cite{Simonyan2015}. 
The CNN convolutional blocks of this CNN is followed by two dense fully connected layers with either 4096 or 1024 neurons.
Its final fully connected layer has either five or 29 neurons, depending on the number of classes.
The training was performed on two fixed-point formats: Q2.14 and Q2.6, and tested on three configurations, 
which will be denoted CNNA$_{16}$, CNNA$_8^1$ and CNNA$_8^2$. 
CNNA$_{16}$ uses the tuning parameters $\vec{\beta} = [16,128,8,3,32]$, 
CNNA$_8^1$ uses $\vec{\beta} = [8,128,16,1,42]$ and CNNA$_8^2$ uses $\vec{\beta} = [8,128,8,3,42]$.

\myfig{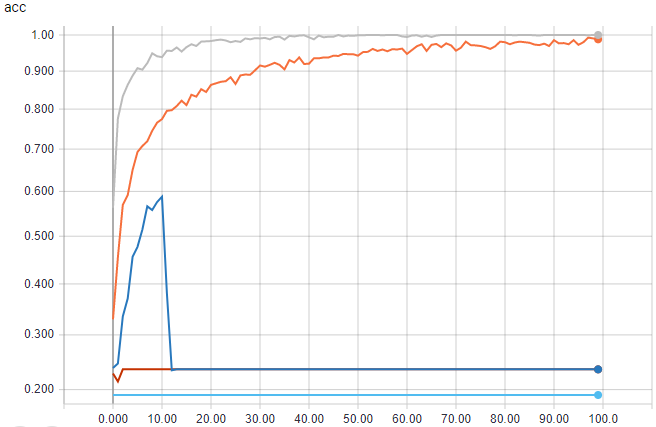}{1.00}
{Training with five classes. Gray: floating-point, orange: fixed-point Q2.14 with auto-scaling, blue: fixed-point Q2.14 without auto-scaling, red at the bottom: fixed-point Q2.6 with and without auto-scale.}

\myfig{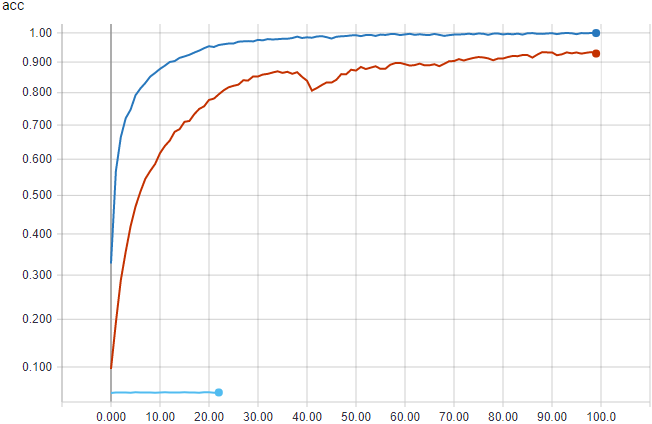}{1.00}
{Training with 29 classes. Blue: floating-point, red: fixed-point Q2.14 with auto-scale, light blue: fixed-point Q2.14 without auto-scaling.}

The accuracy, performance and power consumption of the proposed system will be presented and discussed in this section. 
\subsection{Accuracy}

The CNN was trained using the small dataset in order to find suitable candidates faster, since it is faster to train for five classes than for 29 classes. 
If the accuracy of a fixed-point format is poor on five classes, it will likely be as poor, or worse, when training on 29 classes. 
Therefore, initial training was carried out on the small dataset.

\Figref{detect_small_1024.png} shows that most of the trained models faced issues and obtained low accuracy when using fixed-point format. 
The only quantized version that obtained a high level of accuracy was the one using fixed-point format Q2.14. 
It is unknown why the training with fixed-point format Q2.14 and no auto-scaling makes a sudden dive after 10 epoch. 
However, it could be caused by the learning-rate being too high or too low, or too few neurons in the fully connected layers. 
The best results were achieved with fixed-point format Q2.14 and auto-scaling, which converges towards an accuracy of almost 100\%.
All fixed-point Q2.6 versions did not manage to be trained or achieve any useful results.

Table~\ref{tab:train5results} shows the results of the training with five classes. 
Only the training that used Q2.14 with no auto-scaling performed well with 4096 neurons and reached approximately 83\%. 
The table shows the number of neurons in the fully connected layers, N$_\text{neurons}$, as well as the training and validation accuracy. 
Validation was performed on a dataset not used for training. 

\begin{table}[htbp]
	\centering
	\caption{Training results for the training of VGG16 on five classes.}
	\label{tab:train5results}
	\begin{tabular}{|l|l|l|l|l|}
		\hline
		Type & auto-scale & N$_{neurons}$  & validate & train \\ \hline
		float & $n/a$ & 1024 & $97.5 $ & $ 100.0 $ \\ \hline
		Q2.6 & yes & 1024 & $20.3 $ & $20.9 $ \\ \hline
		Q2.14 & no & 1024 & $24.3 $ & $23.6  $ \\ \hline
		Q2.14 & yes & 1024 & $94.2 $ & $98.8 $ \\ \hline
		float & $n/a$ & 4096 & $97.9 $ & $ 99.5  $ \\ \hline
		Q2.14 & no & 4096 & $83.2 $ & $83.6  $ \\ \hline
		Q2.14 & yes & 4096 & $91.7 $ & $97.6 $ \\ \hline
	\end{tabular}
\end{table}

A final test was performed on all 29 classes of DETECT with, the candidates that performed well in the previous test. 
The best candidates were the floating-point version for reference and the versions that used fixed-point format Q2.14, both with and without auto-scaling. 
As is evident from \figref{detect_full_4096.png}, only the training that used fixed-point format Q2.14 and auto-scaling achieved promising results. 
It shows that it is much more difficult to train the CNNA when using quantization, because details are lost due to the limited range of the fixed point numbers. 
However, it takes many more iterations for the training to reach the same accuracy level as the floating-point format.

The first 29 classes from ImageNet and Cifar-100 were also used for training. 
The validation results in table~\ref{tab:train29results} shows that comparing 
the $Q2.14$ format with floating-point the accuracy drops with $3.5\%$ and $3.2\%$.
For DECTECT the drop is $4\%$ which is higher compared to training with ImageNet and Cifar-100.

\begin{table}[htbp]
	\centering
	\caption{Results for the training of the 16-bits fixed-point VGG16 on 29 classes from the datasets: DETECT(DET), ImageNet(Image) and Cifar-100(Cifar).}
	\label{tab:train29results}
	\begin{tabular}{|l|l|l|l|l|l|}
		\hline
		Type & data & auto & N$_{neurons}$ & val. & train \\ \hline
		float & DET & $n/a$ & 1024 & $88.0 $ & $100.0 $ \\ \hline
		$Q2.14$ & DET & no & 1024 & $5.0 $ & $5.0 $ \\ \hline
		$Q2.14$ & DET & yes & 1024 & $86.4 $ & $94.4 $ \\ \hline
		float & DET & $n/a$ & 4096 & $84.0 $ & $99.4 $ \\ \hline
		$Q2.14$ & DET & no & 4096 & $5.0 $ & $5.1  $ \\ \hline
		$Q2.14$ & DET & yes & 4096 & $86.5 $ & $92.9 $ \\ \hline
		float & Image & $n/a$ & 4096 & $83.0 $ & $99.2 $ \\ \hline
		$Q2.14$ & Image & yes & 4096 & $79.5 $ & $88.1 $ \\ \hline
		float & Cifar & $n/a$ & 4096 & $80.5 $ & $99.5 $ \\ \hline
		$Q2.14$ & Cifar & yes & 4096 & $77.3 $ & $89.3 $ \\ \hline
	\end{tabular}
\end{table}
 
\subsection{Performance}

The Xilinx Ultra96 board~\cite{ultra96} was used to evaluate the performance of the system using a HW clock of 100 MHz and 172.22 MHz for the CNNA IP core. 
The inference time was measured for the different configurations of the CNNA$_{16}$, CNNA$_8^1$ and CNNA$_8^2$  
and the inference times are shown in table \ref{tab:predictionTimeResults}.
The timing performance was measured on the Ultra96 board during inference of the quantized and trained VGG16 model with five classes.
The mean time and variance is an average of 30 measurements. 
The fastest model CNNA$_8^1$ took $1.22$ sec per image, while the slowest, CNNA$_{16}$ at 100 MHz, took $2.20$ sec per image.

\begin{table}[htbp]
	\centering
	\caption{Average inference time and variance using VGG16 for five classes using four different IP cores.}
	\label{tab:predictionTimeResults}
	\begin{tabular}{|l|l|l|l|l|}
		\hline
		           & \begin{tabular}[c]{@{}c@{}} \small CNNA$_{16}$\\{\small 100MHz}\end{tabular}& \begin{tabular}[c]{@{}c@{}} \small CNNA$_{16}$\\{\small 172MHz}\end{tabular} & \begin{tabular}[c]{@{}c@{}} \small CNNA$_8^1$\\{\small 172MHz}\end{tabular} & \begin{tabular}[c]{@{}c@{}} \small CNNA$_8^2$\\{\small 172MHz}\end{tabular}\\ \hline
	avg \tiny $[sec]$ & $2.20$ & $1.96$  & $1.22$  &  $1.49$ \\ \hline
	var \tiny $[\cdot10^{-3}]$ & $0.25$ & $0.30$  & $0.20$  &  $0.11$ \\ \hline
	\end{tabular}
\end{table}

The different layers have different execution times, as shown in table \ref{tab:vgg16layerTimes}. 
As expected, the execution time in convolutional layers depended on the number of bits in the fixed-point format. 
However, pooling took approximately the same time for all tested IP, since pooling is independent of the fixed-point format. 
The table shows that the IP CNNA$_8^1$, obtained the best performance due to the lager number of PEs (16). 
Note that CNNA$_8^2$ was slightly faster than CNNA$_8^1$ in the convolutional layers, even with fewer PEs, due to the higher Bandwidth of the output multiplier.
There is a large number of splits (512) in the dense\_1 and dense\_2 layers, 
and they consume more than half of the total execution time for all three CNNA configurations. 
In average $32\%$ of the time is used to setup the DMA's from PYNQ, which could be optimized with a scatter-gather DMA.
In such a solution the DMA would initiate transfer for the next location of DRAM data without involving the CPU.     
A larger FPGA with more on-chip memory could also be a solution to lower the number of splits and optimize the performance further.

\begin{table}[ht]
	\centering
	\caption{Time of execution of each VGG16 layer in $[ms]$ using four different IP cores.}
	\label{tab:vgg16layerTimes}
	\begin{tabular}{|l|c|c|c|c|}
		\hline
 		layer		&\begin{tabular}[c]{@{}c@{}} \small CNNA$_{16}$\\{\small 100MHz}\end{tabular}& \begin{tabular}[c]{@{}c@{}} \small CNNA$_{16}$\\{\small 172MHz}\end{tabular} & \begin{tabular}[c]{@{}c@{}} \small CNNA$_8^1$\\{\small 172MHz}\end{tabular} & \begin{tabular}[c]{@{}c@{}} \small CNNA$_8^2$\\{\small 172MHz}\end{tabular}\\ \hline
		l1\_conv1 & 19.3 &  17.0   & 19.9 &   21.4     \\ \hline
		l1\_conv2 & 111 &  84.1  & 61.3 &    60.1      \\ \hline
		l1\_pool  & 18.1 &  13.7  & 12.9 &    17.1          \\ \hline
		l2\_conv1 & 55.4 &  42.9  & 31.3 &    30.5       \\ \hline
		l2\_conv2 & 108 &  81.3  & 60.2 &    56.3          \\ \hline
		l2\_pool  & 8.98 &  6.87    & 6.36 &   8.40       \\ \hline
		l3\_conv1 & 56.0 &  43.5  & 33.1 &    29.9       \\ \hline
		l3\_conv2 & 112 &  84.2  & 64.2 &    59.1        \\ \hline
		l3\_conv3 & 110 &  85.5  & 63.0 &    57.8        \\ \hline
		l3\_pool  & 4.51 &  3.48    & 3.25 &   4.23        \\ \hline
		l4\_conv1 & 64.6 &  51.5  & 37.9 &    35.3        \\ \hline
		l4\_conv2 & 126 &  97.9  & 76.0 &    70.5        \\ \hline
		l4\_conv3 & 123 &  102.0  & 73.5 &    67.8        \\ \hline
		l4\_pool  & 2.32 &  1.83  & 1.71 &    2.19        \\ \hline
		l5\_conv1 & 46.6 &  41.0  & 30.7 &    29.3        \\ \hline
		l5\_conv2 & 49.1 &  39.7  & 29.7 &    28.1       \\ \hline
		l5\_conv3 & 45.8 &  39.5  & 29.6 &    27.8   \\ \hline
		l5\_pool  & 0.74 &  0.62  & 0.59 &    0.69   \\ \hline
		dense\_1  & 767 &  737   & 364 &   509   \\ \hline
		dense\_2  & 393 &  397  & 197 &   362   \\ \hline
		dense\_3  & 1.55 &  1.62  & 1.18 &  1.50   \\ \hline        
	\end{tabular}
\end{table}

\subsection{Power consumption}

The power consumption of the design with CNNA$_{16}$, CNNA$_8^1$ and CNNA$_8^2$ was measured on the Ultra96 board 
during inference of the trained VGG16 model with five classes. 
The measured voltage of the power supply to the board was multiplied with the measured current to compute the power consumption. 
The mean and maximum power during inference is calculated as a mean of 10 inferences. 
The power consumption of the IP core is defined as the difference between the Ultra96 board idling and power during inference.
The idle power consumption was measured at $P_\text{idle} = 3.055$ Watt over a five-minute period:

\begin{table}[ht]
	\centering
	\caption{Average and peak power consumption in watt of the Ultra96 board and the IP core during inference.}
	\label{tab:powerConsumption}
	\begin{tabular}{|l|l|l|l|l|}
		\hline
		&\begin{tabular}[c]{@{}c@{}} \small CNNA$_{16}$\\{\small 100MHz}\end{tabular}& \begin{tabular}[c]{@{}c@{}} \small CNNA$_{16}$\\{\small 172MHz}\end{tabular} & \begin{tabular}[c]{@{}c@{}} \small CNNA$_8^1$\\{\small 172MHz}\end{tabular} & \begin{tabular}[c]{@{}c@{}} \small CNNA$_8^2$\\{\small 172MHz}\end{tabular}\\ \hline
		$\text{P}_\text{avg} $ & 5.28 &     5.68  	&   4.71 	& 4.80   	\\ \hline
		$\text{P}_\text{peak}$ 	& 6.60 &     7.14  	&   5.76	& 6.35   	\\ \hline
		$\text{P}_{\text{IP}_\text{avg}}$  & 2.23 &     2.63  	& 1.66 	& 1.74   	\\ \hline
		$\text{P}_{\text{IP}_\text{peak}}$ & 3.55 &    4.09   	&   2.71 	& 3.30 		\\ \hline   
	\end{tabular}
\end{table}

Table~\ref{tab:powerConsumption} shows that the mean power consumption of the Ultra96 board for all tests was between $4.7-5.7$ W 
out of which the IP core only consumes approximately $2$ W. 
This means that running the IP did not affect the average power consumption. 
However, because they run for a shorter amount of time, the fixed-point IPs with a low number of bits used less energy per inference. 
The CNNA$_{16}$ with a $100$ MHz clock was $0.24$ sec slower but consumed less power than the version with a $172$ MHz clock. 
Table~\ref{tab:powerConsumption} shows that the peak power consumption was almost the same for all tested IPs in the range from $2.7$ W to $4.1$ W. 

Figure~\ref{fig:powerConsumptionAll.png} shows that the power consumption is largest in the beginning of the inference, i.e. in the convolution blocks of the CNN. 
The power consumption dropped during execution of the fully connected layers. 
This indicates that most of the FPGA logic was in action during convolution, while less logic was used during computing of the fully connected layers and pooling. 
Pooling activity corresponds to the big dips in power consumption in the first half of the inference.

\myfig{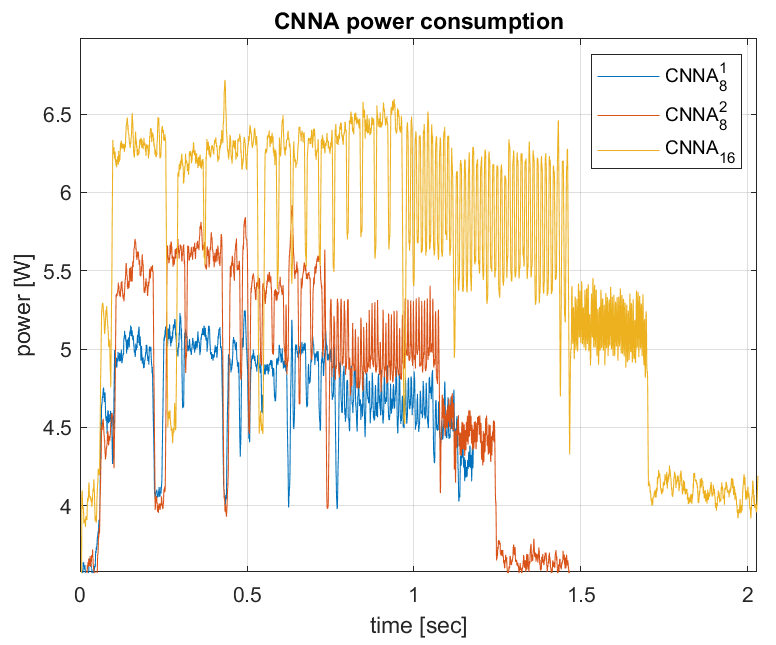}{1.00}
{Power consumption of the tested solution with format CNNA$_8^1$, CNNA$_8^2$ and CNNA$_{16}$ during inference at 172 MHz.}

\section{Comparison with state-of-the-art CNNs}
We have chosen to evaluate our work with the current state-of-the-art toolflows presented in~\cite{Venieris2018} which use a fixed-point resolution of 8 or 16-bits to perform FPGA acceleration of the VGG16 network by targeting the Xilinx Zynq and UltraScale platforms. 
The purpose of this is to compare our work with other tools that have mapped the same CNN on similar FPGA devices from the same vendor. 

\paragraph*{Accuracy} 
Our fixed-point training method only performed well for 16-bit quantization. 
DoReFa-Net~\cite{Shuchang2008} proposes a method for training CNNs with low-bit quantization. 
The method demonstrate a high accuracy bu using AlexNet with only 1-bit weights.
FINN-R~\cite{Blott2018} uses quantized neural networks for low-bit quantization of different CNN topologies with high accuracy.   
Angel-Eye~\cite{Guo2018} also proposes a dynamic quantization strategy, where the network is initially trained with a floating-point format.
The radix position of the fixed-point data is chosen differently for each layer based on statistics, and an optimal radix point is chosen.
The network is converted back to floating-point and fine tuned. 
This method achieves a high level of accuracy for both 16 and 8-bit formats with VGG16.  

\paragraph*{Performance} 
To compare the different solutions, the performance needs to be expressed in Giga Operations Per Second (GOPS) .
The performance result is normalized relative to the number of Look Up Tables (LUTs) and DSPs as a measure for available resources on the target device.
This performance density measure is used to compare the VGG16 mapped to different FPGA devices.
The throughput performance is calculated as the number of Giga Operations (GOP) performed by the CNN relative to the inference time in seconds. 
In the case of the VGG16 network, the total number is $30.76$ GOP out of which $30.7$ GOP is performed in the convolutional (CONV) layers. 
We have presented the performance for CONV layers and all layers of the VGG16 model since some solutions do not accelerate the FC layers. 

\begin{table*}[ht]
	\centering
	\caption{Table comparing the CNNA with state-of-the-art CNN accelerators: DnnWeaver~\cite{Sharma2016}, fpgaConvNet~\cite{Venieris2016}, Angel-Eye~\cite{Guo2018} and Caffeine~\cite{Zhang2016}. All solutions targets the Xilinx Zynq devices except for Caffeine, which uses the Kintex UltraScale FPGA. The power efficiency, performance density and throughput performance are listed for the different solutions. The performance density is only shown for the CONV layers. }
	\label{tab:compare_sota_results}
		\begin{tabular}{|l|l|l|l|l|l|l|l|l|l|}
			\hline
			\multicolumn{1}{|c|}{\textbf{Technique}{\tiny -Mhz}} & \multicolumn{1}{c|}{\textbf{\begin{tabular}[c]{@{}c@{}}Power \\ Efficiency\\ {\tiny [GOPS/W]}\end{tabular}}} & \multicolumn{1}{c|}{\textbf{\begin{tabular}[c]{@{}c@{}}Power E.\\(Conv)\\{\tiny [GOPS/W]}\end{tabular}}} & \multicolumn{1}{c|}{\textbf{\begin{tabular}[c]{@{}c@{}}Density\\ {\tiny [GOPS/} \\ {\tiny DSP]}\end{tabular}}} & \multicolumn{1}{c|}{\textbf{\begin{tabular}[c]{@{}c@{}}Density\\ {\tiny [GOPS/} \\ {\tiny kLUT]}\end{tabular}}} & \multicolumn{1}{c|}{\textbf{\begin{tabular}[c]{@{}c@{}}Perfor.\\ (Conv)\\ {\tiny [GOPS]}\end{tabular}}} & \multicolumn{1}{c|}{\textbf{\begin{tabular}[c]{@{}c@{}}Perfor.\\(Total)\\ {\tiny [GOPS]}\end{tabular}}} & \multicolumn{1}{c|}{\textbf{\begin{tabular}[c]{@{}c@{}}Xilinx\\ Device\end{tabular}}} & \multicolumn{1}{c|}{\textbf{\begin{tabular}[c]{@{}c@{}}Fix.\\ {\tiny [bits]}\end{tabular}}} \\ \hline
			\textbf{\scriptsize DnnWeaver}{\tiny -150}    & $n/a$   & $n/a$   & $0.143$  & $0.59$ & $31.4$ & $n/a$  & ZC7Z020 & 16 \\ \hline
			\textbf{\scriptsize fpgaConvNet}{\tiny -125}  & $n/a$   & $7.27$  & $0.221$  & $0.91$ & $48.5$ & $12.7$ & ZC7Z020 & 16 \\ \hline
			\textbf{\scriptsize fpgaConvNet}{\tiny -125}  & $n/a$   & $n/a$   & $0.173$  & $0.71$ & $156$  & $n/a$  & ZC7Z045 & 16 \\ \hline
			\textbf{\scriptsize Angel-Eye}{\tiny -214}    & $n/a$   & $24.1$  & $n/a$    & $n/a$  & $85.3$ & $n/a$  & ZC7Z020 & 8  \\ \hline
			\textbf{\scriptsize Angel-Eye}{\tiny -150}    & $14.2$  & $n/a$   & $0.209$  & $0.86$ & $188$  & $137$  & ZC7Z045 & 16 \\ \hline
			\textbf{\scriptsize Caffeine}{\tiny -200}     & $10.64$ & $12.4$  & $0.187$  & $1.55$ & $310$  & $266$  & KU060   & 16 \\ \hline
			\textbf{\scriptsize CNNA$_{16}$}{\tiny -100}  & $6.28$  & $11.86$ & $0.078$  & $0.62$ & $26.4$ & $14.0$ & ZU3EG   & 16 \\ \hline
			\textbf{\scriptsize CNNA$_{16}$}{\tiny -172}  & $5.99$  & $11.08$ & $0.081$  & $0.52$ & $29.1$ & $15.7$ & ZU3EG   & 16 \\ \hline
			\textbf{\scriptsize CNNA$_{8}^1$}{\tiny -172} & $15.22$ & $22.94$ & $0.155$  & $0.66$ & $38.0$ & $25.2$ & ZU3EG   & 8  \\ \hline
			\textbf{\scriptsize CNNA$_{8}^2$}{\tiny -172} & $11.83$ & $22.53$ & $0.110$  & $0.61$ & $39.5$ & $20.7$ & ZU3EG   & 8  \\ \hline
		\end{tabular}
\end{table*}

The results are shown in table~\ref{tab:compare_sota_results}, which indicates that the CNNA performance (Total) 
is lower than the comparable state-of-the-art architectures. 
The best performance of our 16-bit solution is 29.1 GOPS.
This is lower than the 31.4 GOPS for D{\footnotesize NN}W{\footnotesize EAVER} which has the worst performance of the state-of-the-art solutions.

The Ultra96 target used in our evaluation is small and low-cost compared to the ones used in some of the examples e.g. Zynq XC7Z2045 and UltraScale KU060. 
If a larger and more expensive target such as the Xilinx ZCU104 evaluation kit~\cite{zcu104} was used, 
it would be possible to increase the number of PEs, thereby achieving a higher throughput and performance.

The performance density measure is also lower than most the other architectures and only similar to D{\footnotesize NN}W{\footnotesize EAVER}.
Angel-Eye and Caffeine both have a much higher density performance compared to usages of LUT and DSP resources on the FPGA.  

\paragraph*{Power efficiency}
The power efficiency is dependent on both of the efficiency of data communication and computation. 

The SmartShuttle~\cite{Li2018} solution is optimizing CNN off-chip memory access.
Observing that over $80\%$ of energy is consumed by DRAM accesses,
they make a benchmark of the data volume of DMA requests during inference of the 13 CONV layers in VGG16. 
Our CNNA16 measures a data volume of 211.7 MB transferred for the same feature layers including pooling. 
However, we use more on-chip memory for weight and data buffers than SmartShuttle.  
As a benchmark SmartShuttle measures 221.3 MB. 
Simulated with a on-chip buffer of 512 KB, however, they can lower the DRAM access volume to 160 MB. 
The design of the CLB in our CNNA ensures that weights are only transferred once from DRAM,
which is similar to what SmartShuttle achieves with the weight reuse oriented scheme (WRO) they propose.
The last three FC layers of the CNNA16 transfers a volume of 273.8 MB, 
which is not considered by SmartShuttle and stands for most of the data communication.
  
The computation power efficiency is calculated as the number of operations per second, 
relative to the the mean power consumption of the CNNA, which we measured earlier (GOPS/W). 	
Compared to many of the current state-of-the-art accelerators, the CNN accelerator in this work performs quite well in terms of power efficiency. 
When using 16-bit fixed-point weights at 100 MHz, its total power efficiency is 0.44x lower than Angel-Eye and 0.59x lower than Caffeine. 
With nearly the same efficiency of 12 GOPS/W, the power efficiency of the CONV layers are considered comparable with Caffeine. 
The performance bottleneck in our CNN accelerator is the fully connected layers, where splits are performed 512 times with a high DRAM access. 
The fpgaConvNet on the Zynq XC7Z020 has a worse efficiency of 7.3 GOPS/W compared to the CNNA16 with 11.9 GOPS/W.
While Angel-Eye's fixed-point 8-bit with 24.1 GOPS/W is the best of all the compared state-of-the-art solutions in terms of efficiency, 
the 8-bit CNNA with 23.0 GOPS/W is a close second.
 
\section{Conclusion}

In this paper, an architecture for a SoC design was presented.
The presented architecture implements the different operations necessary 
for a deep neural network to perform close to real-time inference. 
The architecture was implemented using Python and HLS for the IP core and was able to run on the Ultra96 board using PYNQ. 
The interface for the system is similar to Keras and should be familiar to most engineers working in the field of machine learning. 

The CNN is able to accelerate deep learning algorithms that use any sequence of convolutional, max-pooling and fully connected layers. 
The layer operations can support many different parameters and will be able to perform inferences using most modern CNNs. 
The network weights can use any 8-, 16- or 32-bit fixed-point format when exported from Keras with the weights auto-scaled correctly. 
A training method was proposed which achieved high levels of inference accuracies, both when using fixed-point and floating-point weights.
The VGG16 architecture chosen for testing in this paper was able to perform inference in $2.0$ sec per image 
when using the fixed-point format Q2.14 and $1.2$ sec when using fixed-point format Q2.6. 
The IP core alone consumes a peak power of $4.1$ W with a mean power between $1.5-2.7$ W
and has a power efficiency between $6.0-15.2$ GOPS/W depending of the fixed-point format.

Compared to similar state-of-the-art solutions for mapping the VGG16 network to Xilinx platforms, our solution demonstrates a comparable energy efficiency, especially for the convolutional layers.
In future work, the CNNA needs be extended to support special layers to support deep neural networks such as ResNet, DenseNet, InceptionNet and GooglLeNet.
The special layers with irregular dataflow will be implemented in the SW controlling part of the proposed architecture.
	
\section*{Acknowledgments}

We would like to thank Freia Martensen for language and proof reading the article.

\bibliography{GenericCNNA}

\end{document}